\documentclass[letterpaper, 10 pt, conference]{ieeeconf}  

\IEEEoverridecommandlockouts                              
\usepackage{amsmath}
\usepackage{amssymb}
\usepackage{siunitx}
\usepackage[colorlinks,linkcolor=magenta]{hyperref}
\usepackage{float}
\usepackage{multirow}

\usepackage{cite}
\usepackage{graphicx}
\usepackage{subcaption}

\newcommand{\etal}{\textit{et al}. }
\newcommand{\norm}[1]{\left\lVert#1\right\rVert}

\usepackage{xcolor}
\usepackage{soul}

\newcommand\rurl[1]{%
  \href{http://#1}{\nolinkurl{#1}}%
}

\DeclareMathOperator*{\argmin}{arg\,min}

\title{\LARGE \bf
Level Set-Based Camera Pose Estimation From Multiple 2D/3D Ellipse-Ellipsoid Correspondences
}

\author{Matthieu Zins, Gilles Simon, Marie-Odile Berger
\thanks{This work was supported by the MoveOn project (Inria - DFKI).}
\thanks{Authors are with Inria, Université de Lorraine, LORIA, CNRS.\newline
        {\tt\small \; forename.name@inria.fr}}%
}

\begin{document}

\maketitle
\thispagestyle{empty}
\pagestyle{empty}

\begin{abstract}


In this paper, we propose an object-based camera pose estimation from a single RGB image and a pre-built map of objects, represented with ellipsoidal models.  We show that contrary to point correspondences, the definition of a cost function characterizing the projection of a 3D object onto a 2D object detection is not straightforward. We develop an ellipse-ellipse cost based on level sets sampling, demonstrate its nice properties for handling partially visible objects and compare its performance with other common metrics. Finally, we show that the use of a predictive uncertainty on the detected ellipses allows a fair weighting of the contribution of the correspondences which improves the computed pose. The code is released at {\small \rurl{gitlab.inria.fr/tangram/level-set-based-camera-pose-estimation}}.
\end{abstract}

\section{Introduction}





Estimating the 6-DoF pose of a camera from an RGB image is a fundamental task in computer vision, with application in robotics, autonomous systems or Augmented Reality.

The performance of classical point-based methods depends mostly on the ability to perform correct 2D-3D matching. While these methods can achieve very good accuracy, they require heavy computations to handle outliers and large point clouds as scene models. Such point clouds generally do not contain semantic information and lack of interpretability. Also, they are limited to highly textured environments.


Recently, object-based camera pose estimation has become an attractive research direction thanks to the robustness in object detection and recognition provided by deep learning.
Objects have been integrated in Simultaneous Localization and Mapping (SLAM) using different modellings, such as cuboids~\cite{cubeslam} or ellipsoids~\cite{quadricslam}. However, these methods still rely on point-based tracking or odometry measurements. SLAM++~\cite{slam++} and DeepSLAM++~\cite{deepslam++} combined an object detector with CAD models for object-level \hbox{RGB-D} SLAM, but necessitate depth measurements as well as precise 3D models of the objects to train their networks.

\begin{figure}[ht]
    \centering
    \includegraphics[width=0.92\linewidth]{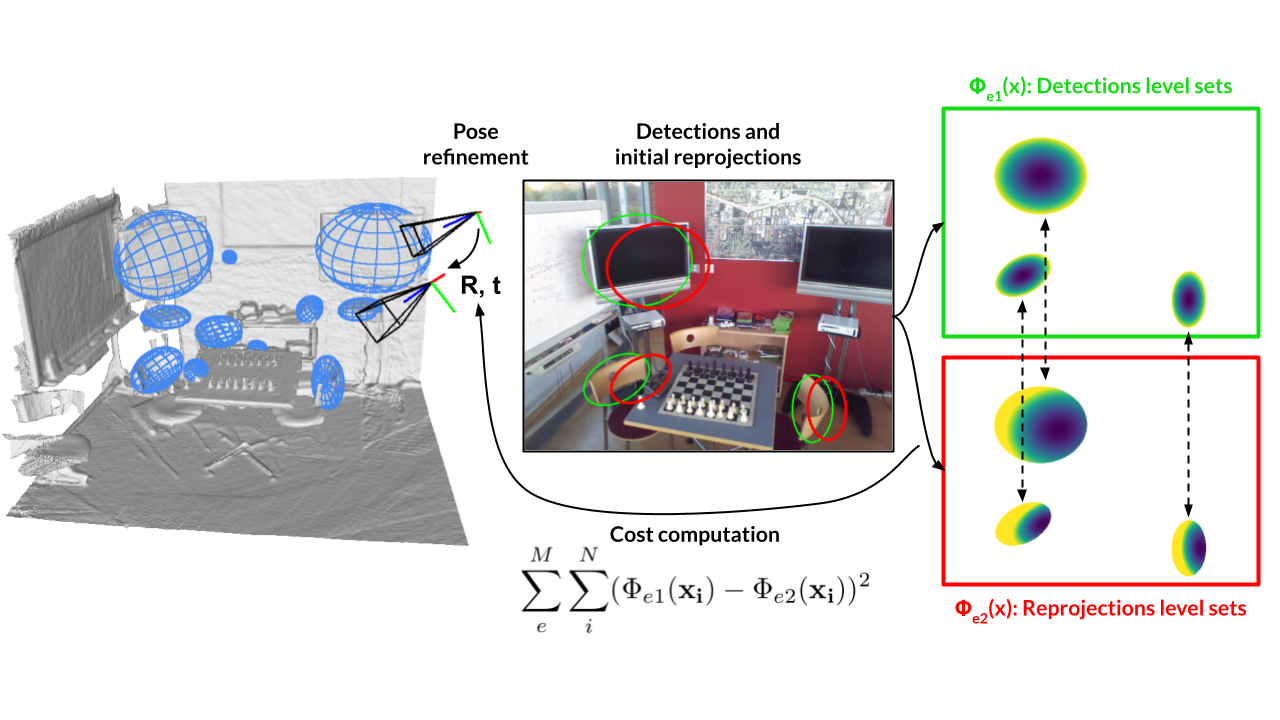}
    \caption{Camera pose estimation by minimizing the reprojection error expressed with a level set-based metric.}
    \label{fig:pose_refinement_level_sets}
\end{figure}

In this work, we are interested in relocalizing the camera from a single RGB image with respect to a pre-built map of objects, which is of particular interest in case of tracking failure or for re-initializing SLAM in a previously scanned environment (e.g., kidnapping problem).
We place ourselves in the challenging, but more realistic, context where we do not have access to precise 3D models of the objects in the scene, and instead, we use simplified ellipsoidal models.


Close to our work,~\cite{3DV} also models objects with ellipsoids and proposes improved elliptic detections to compute an initial camera pose. However, this is more a rough estimate of the camera pose which is computed from only a minimal set of two or three objects and the additional objects are used in a validation step in order to select the best solution. Also, its accuracy is limited by two geometric approximations: the camera roll is null and the center of an ellipsoid projects into the center of the ellipse. In this work, we go one step further and propose a pose estimation method which optimizes ellipses alignment between the 2D object detections and the projection of their 3D ellipsoidal models, taking into account all the detected and matched objects.

\begin{figure}
    \centering
    \includegraphics[width=\linewidth]{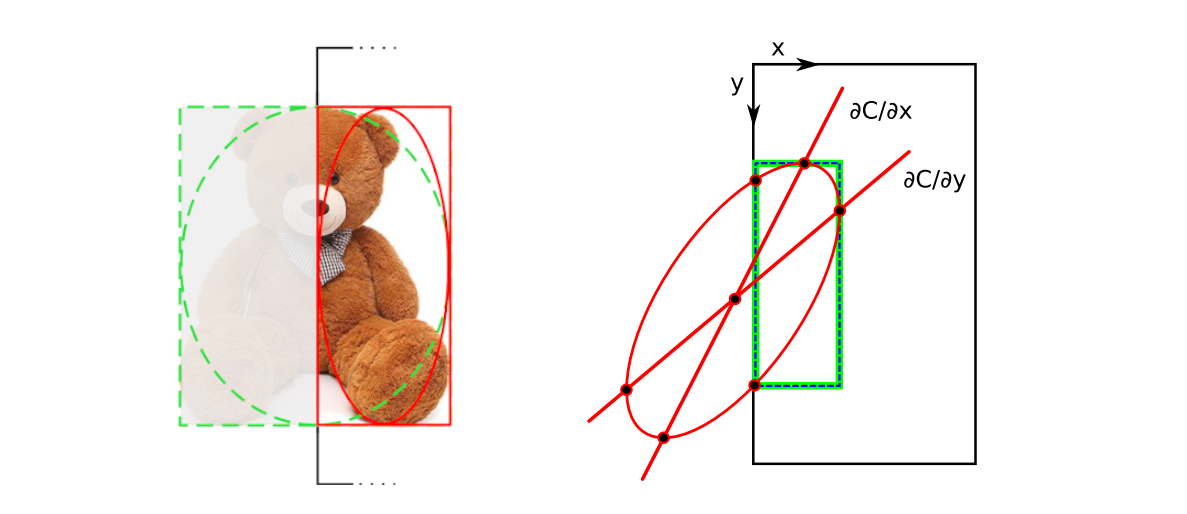}
    \caption{Left: Incorrect detection (red) of an object partially outside the image. In reality, the detection of a partially visible object is usually even worse (see the bottom chair in Figure~\ref{fig:results_chess_frames_comparison_frames_2090}). Right: QuadricSlam bounding box~\cite{quadricslam}, restricted to the on-image ellipse.}
    \label{fig:partially_visible_objects_explain}
\end{figure}

One of the major challenges in estimating the camera pose from objects comes from partially visible objects. These can be occluded by other elements of the scene or partially outside the image, which, for example, happens frequently when the camera is moving. The detection of such objects are generally of poor quality and limited to the visible parts (see Figure~\ref{fig:partially_visible_objects_explain}). 
Also, contrary to keypoints which are generally in great number in an image, there are usually much less detected objects and it is therefore not conceivable to completely discard the poorly detected ones, provided that we are even able to identify them. Instead, a smart balancing of their contribution in the pose is required.

In QuadricSlam~\cite{quadricslam}, Nicholson~\etal restrict the projection of the ellipsoidal model to the on-image conic (i.e. the part of the ellipse which is inside the image, see Figure~\ref{fig:partially_visible_objects_explain}). This helps to deal with objects whose undetected parts are actually outside the image. 
However, objects that are entirely inside the image may also be poorly detected, because of occlusion or simply due to an uncommon viewoint which challenges the object detection network.

We compare different ellipse-ellipse metrics and show that our proposed cost based on level sets has nice properties for handling partial detections.
Also, inspired by the recent advances in estimating the uncertainty of a neural network prediction, we propose a weighting method that can help in determining which objects should be trusted and which should be given little importance.


Our contributions are the following:
\begin{itemize}
    \item We develop an optimization-based method for camera pose estimation from multiple objects observations.
    \item We propose to use level sets to establish a metric between ellipses, and demonstrate its nice properties for dealing with poor-quality object detections.
    \item We show that integrating a confidence measure in the pose refinement improves the obtained pose accuracy, especially in case of partially visible objects.
\end{itemize}


\section{Related work}
\subsection{Camera pose estimation}

Absolute camera pose estimation was traditionally addressed by sequentially solving two sub-problems: first a feature matching problem that seeks to establish putative 2D-3D correspondences, and then a Perspective-n-Point problem~\cite{epnp} that minimizes a reprojection error with respect to the camera pose. With the advances of deep learning, some parts have been replaced by learned components, such as Superpoint~\cite{superpoint} for local features detection and description or Superglue~\cite{superglue} for matching. Recently, detector-free matching, which directly produces dense matches without the detection phase, have also shown promising results~\cite{loftr}.

End-to-end learning of absolute camera pose has also gained attention with PoseNet and its evolutions~\cite{posenet, poselstm}. While these methods provide an interesting robustness to illumination changes or motion blur, they do not reach the same level of accuracy as the classical structure-based methods.
Scene coordinates regression has also been explored for camera pose estimation, originally with random forests~\cite{Shotton_forest}, and later with CNNs~\cite{BrachmannR18, BuiAIN18}. These methods provide accurate localization, but are limited to small environments due to the fixed capacity of the network and usually require depth information for training. Most of these learning-based methods require a per-scene training and only very recent works achieve generalization across scenes~\cite{back_to_the_feature, sanet}.

With the emergence of deep learning and the significant advances in object detectors such as Yolo~\cite{yolov4} or Faster R-CNN~\cite{FasterRCNN}, object-based localization methods have become of particular interest. Weinzaepfel~\etal proposed to use planar objects as object-of-interest for visual localization~\cite{planar-object-of-interest}. They predict dense 2D-2D correspondences between the detected object in the query image and a reference image of this object for which 3D coordinates are known.
Labbé~\etal~\cite{cosypose} proposed an object-level bundle adjustment to refine both the poses of cameras and objects in the context of multi-cameras and multi-objects. However, they assume to have access to the 3D models of the objects, which is generally not possible in the context of SLAM.
Yang~\etal developed a monocular object SLAM in which they represent objects with cuboids~\cite{cubeslam}. They infer initial cuboids proposals from 2D bounding boxes and refine them in a bundle-adjustment, simultaneously with the camera poses and points.

Representing objects with 3D cuboids and their observations in images with 2D bounding boxes does not allow to derive closed-form solutions to the projection equations and leads to solutions with a high combinatorics to match the 3D box corners with the 2D box edges. Li~\etal reduced this complexity by training a viewpoint classifier~\cite{Li}. Another solution is to model 2D/3D objects with ellipses/ellipsoids. Crocco and Rubino leveraged this representation in the context of 3D object localization~\cite{rubino_ellipsoid_reconstruction} and Structure-from-Motion with objects~\cite{crocco_sfm_with_objects, psfmo}, using a simplified camera model.
Nicholson~\etal integrated this object representation in QuadricSLAM~\cite{quadricslam}, where the problem is expressed in the form of a factor graph, combining odometry and objects measurements. Similarly, Hosseinzadeh~\etal combined points, planes and quadrics in~\cite{quadrics_and_planes_slam}.


While the previously mentioned works start from an initial pose estimate and are more related to bundle adjustement combining points, objects, planes and odometry measurements, recent works have proposed direct solutions for object-based camera pose estimation.
Gaudilliere~\etal developed a method to estimate the camera pose from two pairs of ellipse/ellipsoid~\cite{gaudilliere}. Zins~\etal extended this work with improved elliptic detections of objects~\cite{3DV}, replacing the original axis-aligned ellipses directly inferred from the bounding boxes provided by the object detector.
In these methods, data association is solved simultaneously with the pose, within a RANSAC loop. However, only two or three objects actually contribute to the estimated pose and its accuracy may be limited by two assumptions: the camera roll is null and the center of an ellipsoid projects into the center of the projection ellipse. A refinement step, based on the maximization of Intersection-over-Union between ellipses, was proposed in~\cite{gaudilliere} in order to consider all the detected objects, but without success, as it did not improve the pose accuracy and even deteriorated it in some cases.

\subsection{Uncertainty estimation}

In the last years, many works attempted to quantify the uncertainty of neural networks predictions, in particular with Bayesian neural networks \cite{ABDAR2021243}. 
In~\cite{what_uncertainties}, Gal and Kendall distinguished two types of uncertainties: the aleatoric uncertainty which captures the noise in the observations and the epistemic uncertainty which represents the uncertainty in the model and which could be reduced with more training data.
They proposed to estimate the aleatoric uncertainty in regression tasks using an observation noise parameter which can be interpreted as a learned loss attenuation. Furthermore, they introduced Monte-Carlo Dropout as a Bayesian approximation of the model weights posterior distribution in order to estimate the epistemic uncertainty. However, several forward passes are required during inference which limits its applicability. Also, adding dropout layers to a neural network raises the questions of where should they be inserted, as well as, what dropout ratio should be used. These values are usually arbitrarily chosen, but can have a significant impact on the predicted uncertainty values.
Recently, ensemble methods have been advised as the new go-to method for estimating the epistemic uncertainty~\cite{ensembles_uncertainty}, but again, their computational cost is high, with multiple forward passes at inference and multiple network trainings.

In this work, we follow a pragmatic approach and focus on estimating the aleatoric uncertainty in order to weight the contribution of the objects to the pose. In particular, we aim at decreasing the influence of badly detected objects.

\textbf{Outline:} In section~\ref{sec:ellipse_ellipse_metrics}, we review different ellipse-ellipse metrics and propose to use a cost based on level sets.
We then compare the metrics on a 2D ellipse registration problem in section~\ref{sec:experiments_ellipses_alignment}. Finally, our 6-DoF camera pose estimation method is explained in section~\ref{sec:camera_pose_refinement} and evaluated in section~\ref{sec:experiments_camera_pose_refinement}, as well as the chosen minimization metrics and the predictive uncertainty. The experiments of sections~\ref{sec:experiments_ellipses_alignment} and~\ref{sec:experiments_camera_pose_refinement} are also illustrated in the accompanying video.

\section{Ellipse-ellipse metrics}
\label{sec:ellipse_ellipse_metrics}


\paragraph{Intersection-over-Union}
It is a widely-used metric for comparing shapes, with applications in semantic segmentation~\cite{cityscapes}, object detection~\cite{coco} and tracking~\cite{vot2016}.
The Jaccard distance is often minimized (e.g., in~\cite{gaudilliere}), such that
\begin{equation}
\Delta_{IoU}(\mathcal{E}_1, \mathcal{E}_2) = 1 - \frac{|\mathcal{E}_1 \cap \mathcal{E}_2|}{|\mathcal{E}_1 \cup \mathcal{E}_2|}.
\end{equation}

A major weakness of IoU is that it remains constant when the two ellipses are disjoint, and thus, equally quantifies them, independently of their distance. Obviously, this behaviour is not desirable for an optimization problem.
\hbox{Rezatofighi~\etal} address this issue in~\cite{giou}, where they propose a generalized version of IoU, abbreviated GIoU, and given by
\begin{equation}
\Delta_{GIoU}(\mathcal{E}_1, \mathcal{E}_2) = 1 -\left(IoU-\frac{|C\setminus(\mathcal{E}_1 \cup \mathcal{E}_2)|}{|C|}\right),
\end{equation}
where $C$ is the smallest convex object enclosing the ellipses.



\paragraph{Bounding box corners}

This metric is defined between the axis-aligned bounding boxes of the two ellipses and is computed as the quadratic error between their corners coordinates, such that

\begin{equation}
\Delta_{bbox}(\mathcal{E}_1, \mathcal{E}_2) = \norm{\mathcal{B}_1 - \mathcal{B}_2}_2^2,
\end{equation}
where $\mathcal{B}_i$ contains the bounds of the i-th box ($min_x$, $min_y$, $max_x$, $max_y$). The main issue of this metric is that an axis-aligned bounding box does not define a unique ellipse, and thus, an infinite number of ellipses share the same bounding box. This does not guarantee ellipses to be aligned even if their bounding boxes perfectly match. Also, boxes depend on the image axes, which induces that rotated pairs of ellipses can have different distances. QuadricSlam~\cite{quadricslam} uses an improved version of this metric, abbreviated QBbox in the next sections, to better handle objects partially outside the image (see Figure~\ref{fig:partially_visible_objects_explain}).


\paragraph{Algebraic error}
The ellipses can also be compared using their dual-form matrices. Such a matrix represents an ellipse by the envelope of all the tangent lines to its curve.
In the context of 3D ellipsoid estimation from multi-view elliptic observations, Rubino~\etal~\cite{rubino_ellipsoid_reconstruction} minimized the quadratic error between vectorized versions of these dual-form matrices $\mathbf{C}^*_{1|2}=\mathbf{C}^{-1}_{1|2}$, given by 
\begin{equation}
    \Delta_{vec}(\mathcal{E}_1, \mathcal{E}_2) = \norm{vec(\mathbf{C}_1^*) - vec(\mathbf{C}_2^*)}^2_2,
\end{equation}
where $vec(\cdot)$ extracts the five upper elements of a matrix.

Instead, the \textit{Frobenius} norm of the difference between the matrices is used in~\cite{quadrics_and_planes_slam}, expressed as
\begin{equation}
    \Delta_{fro}(\mathcal{E}_1, \mathcal{E}_2) = \sqrt{Tr((\mathbf{C}_1^* - \mathbf{C}_2^*)(\mathbf{C}_1^* - \mathbf{C}_2^*)^T)},
\end{equation}
where $Tr(.)$ is the trace.

Both of these metrics, the \textit{vectorized} version and the \textit{Frobenius} one, actually compare the ellipses contours. While this can provide a good accuracy, considering only the contours may suffer from a limited robustness when the two initial ellipses differ a lot. 
Also, the values of these metrics change if a global translation is applied on the two ellipses, which is not desirable when several pairs of ellipses, spread over the entire image, are jointly optimized. Finally, it is difficult to have a geometric interpretation of such distances based on the dual-form matrices of ellipses.

\paragraph{Distribution-based distances}

Another kind of metric that can be used to compare ellipses is based on probability distributions. Indeed, an ellipse, parametrized by two axes ($\alpha, \beta$), an orientation ($\theta$) and a center ($c_x, c_y$), can be interpreted as a 2D Gaussian distribution $\mathcal{N}(\mu,\,\Sigma)$ where

\begin{equation}
        \mu = \left[\begin{array}{c}
            c_x \\
            c_y 
        \end{array}\right], \;\;
        \Sigma^{-1} = R(\theta) \left[\begin{array}{cc}
            \frac{1}{\alpha^2} & 0 \\
            0 & \frac{1}{\beta^2}
        \end{array}\right] R(\theta)^T.
\end{equation}


The \textit{Wasserstein} distance, also called \textit{Earth-moving} distance, can be used to compare such distributions. Intuitively, it represents the cost of transforming one distribution into another one. Its good convergence properties have been demonstrated by Arjovsky~\etal in~\cite{wasserstein_GANs} for distribution learning in the context of GANs. It has also been used as a loss for training an ellipse detection network in~\cite{knots_detection}. While this distance is difficult to compute in general, a closed-form formula exists for the 2D Gaussian case, given by
\begin{equation}
\label{equ:Wasserstein}
\begin{split}
    \Delta_{\mathcal{W}_2^2}(\mathcal{N}_1, \mathcal{N}_2) = & \norm{\mu_1-\mu_2}_2^2 \\
    & + Tr(\Sigma_1 + \Sigma_2 - 2 (\Sigma_1^{\frac{1}{2}} \Sigma_2 \Sigma_1^{\frac{1}{2}})^{\frac{1}{2}}).
\end{split}
\end{equation}

Finally, the \textit{Bhattacharyya}~\cite{bhatta} distance also expresses a measure of similarity between probability distributions.
In the case of two 2D Gaussian distributions, it is calculated by
\begin{equation}
\begin{split}
    \Delta_\mathcal{B}(\mathcal{N}_1, \mathcal{N}_2) = &\frac{1}{8} (\mu_1 - \mu_2)^T \Sigma^{-1}(\mu_1 - \mu_2) \\ & + \frac{1}{2} \ln \left( \frac{\det \Sigma}{\sqrt{\det \Sigma_1 \det \Sigma_2}}\right),
\end{split}
\end{equation}
where $\Sigma=\frac{\Sigma_1 + \Sigma_2}{2}$.


\section{Level set-based metric}


In this section we propose another metric based on level sets and implicit representation of ellipses~\cite{level-sets}, which combines the advantages of considering the contours (accuracy) and the area (better robustness).

\begin{figure}[ht]
    \centering
    \includegraphics[width=0.4\linewidth]{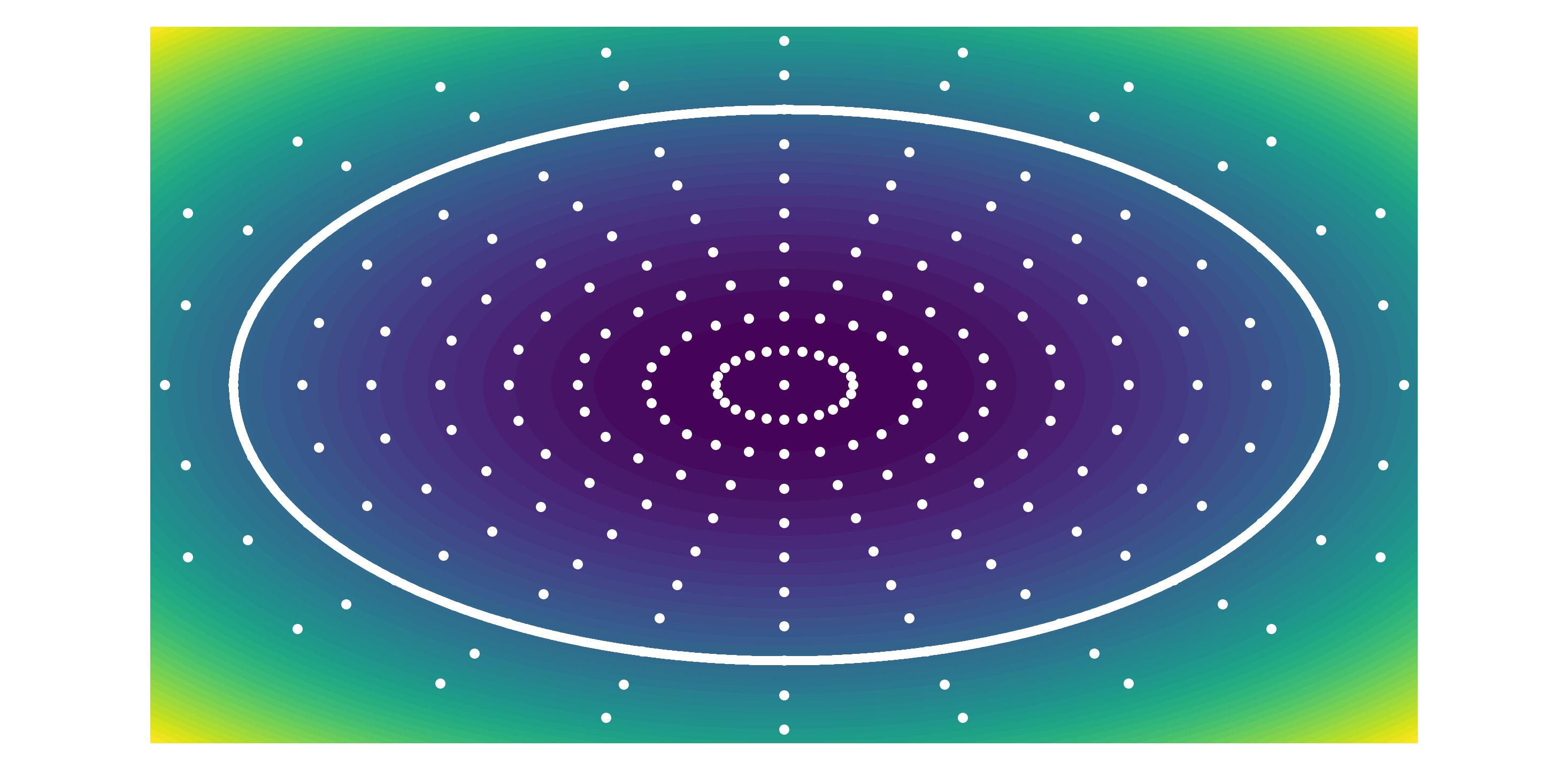}
    \caption{Embedding function $\Phi$ and sampling points.}
    \label{fig:ellipse_level_sets}
\end{figure}

For that, we represent an ellipse with an embedding function $\Phi: \mathbb{R}^2 \rightarrow \mathbb{R}$. We use the general quadratic equation of an ellipse, where its contour corresponds to the level curve of value 1, expressed as
\begin{equation}
    \Phi(\mathbf{x}) =  (\mathbf{x} - c)^T R(\theta) \left[\begin{array}{cc}
        \frac{1}{\alpha^2} & 0 \\
        0 & \frac{1}{\beta^2}
    \end{array}\right] R(\theta)^T (\mathbf{x} - c).
\end{equation}
Then, the level set-based distance between two ellipses represented by their embedding functions $\Phi_1$ and $\Phi_2$ is defined by
\begin{equation} 
    \Delta_\Omega(\Phi_1, \Phi_2) = \int_\Omega (\Phi_1(\mathbf{x})-\Phi_2(\mathbf{x}))^2 \mathbf{dx}.
\end{equation}
To keep the computational cost low, the distance is calculated on a finite set of points, regularly sampled along the level curves of the first ellipse (see Figures~\ref{fig:pose_refinement_level_sets} and~\ref{fig:ellipse_level_sets}), such that
\begin{equation}
    \Delta_{lvs}(\mathcal{E}_1, \mathcal{E}_2) = \sum_{i=1}^N (\Phi_1(\mathbf{x_i})-\Phi_2(\mathbf{x_i}))^2,
\end{equation}
where $\mathbf{x_i}$ is a sample point and $N$ the total number of points.
The size of the sampling area can be adapted depending on the original distance between the two ellipses to ensure a better convergence. Also, in contrast to sampling points on a rectangular grid aligned with the image axes, our sampling preserves the invariance to rotation.


\section{Ellipse 2D registration}
\label{sec:experiments_ellipses_alignment}




We first evaluate each metric on a reduced 2D pose estimation problem, which consists of finding the rotation and translation between two ellipses $\mathcal{E}$ and $\mathcal{E}_{ref}$. $\mathcal{E}_{ref}$ is randomly generated and transformed into $\mathcal{E}$ with a rotation ([$-180^{\circ}, 180^{\circ}$]) and a translation ($[-60, 60]$ pixels). A detection noise (random anisotropic scaling, uniformly sampled between 0.83 and 1.2) is added to $\mathcal{E}$ to simulate an imperfect detection. The goal is to re-estimate the rigid transform by minimizing a cost between the ellipses, expressed as

\begin{equation}
    \hat{\theta},\hat{t} = \argmin_{\theta,t} \Delta(\mathcal{T}(\mathcal{E}, \theta, t), \mathcal{E}_{ref}),
\end{equation}
where $\mathcal{T}(\mathcal{E}, \theta, t)$ denotes rotating ellipse $\mathcal{E}$ by $\theta$ and translating it by $t$.
While this experiment only involves a reduced problem compared to our targeted application of camera pose estimation, it highlights different behaviors of the metrics during the optimization.
In Table~\ref{tab:ellipse_pose_2d}, we can notice that the generalized IoU as well as the algebraic distances have troubles to converge. For example, the optimization gets stuck with a flat cost using the generalized IoU when one ellipse is entirely inside the other one.
On the contrary, the probabilistic metrics and the level set one perform the best. Notice that the level set metric is not subject to the flat cost problem encountered by the generalized IoU, thanks to the form of the chosen embedding function.

\begin{table}
\begin{center}
\begin{tabular}{|c|cc|cc|}
\hline
\multirow{2}{*}{Metric} & \multicolumn{2}{c|}{No noise} & \multicolumn{2}{c|}{With noise} \\
 & \multicolumn{1}{c}{Pos. err.} & \multicolumn{1}{c|}{Rot. err.} & \multicolumn{1}{c}{Pos. err.} & \multicolumn{1}{c|}{Rot. err.} \\
\hline 
Generalized IoU & 13.32 & 6.42 & 15.11 & 12.44 \\
Bounding box & $1.0\mathrm{e}^{-7}$ & 22.87 & $1.2\mathrm{e}^{-5}$ & 29.66 \\
Algebraic Vectorized & 2.49 & 21.69 & 5.05 & 32.30 \\
Algebraic Frobenius & 2.59 & 22.08 & 5.04 & 32.30 \\
Wasserstein & $9.8\mathrm{e}^{-4}$  &  $7.8\mathrm{e}^{-4}$ &  $7.9\mathrm{e}^{-4}$ &  $4.6\mathrm{e}^{-4}$ \\
Bhattacharyya &  $2.0\mathrm{e}^{-2}$ &  $3.9\mathrm{e}^{-3}$ &  $1.7\mathrm{e}^{-2}$ &  $2.5\mathrm{e}^{-3}$ \\
Level Set &  $1.0\mathrm{e}^{-6}$ &  $1.0\mathrm{e}^{-6}$ &  $2.9\mathrm{e}^{-4}$ &  $2.7\mathrm{e}^{-5}$ \\
\hline
\end{tabular}
\end{center}
\caption{Mean position (pixels) and rotation (degrees) errors obtained on the estimated transform for 10000 pairs.}
\label{tab:ellipse_pose_2d}
\end{table}

\section{Camera pose refinement}
\label{sec:camera_pose_refinement}
The previously described ellipse-ellipse distances are the main element of the proposed object-based camera pose estimation, as our ellipsoidal models of objects are observed in the image in the form of ellipses. The projection ellipse can be computed by
\begin{equation}
    C^* = PQ^*P^T,
\end{equation} where $P$ is the camera projection matrix, $Q^*$ is the dual-form matrix of the ellipsoid and $C^*$ is the dual-form matrix of the ellipse.

\subsection{Non-linear minimization problem}
 Similarly to the traditional structure-based methods, where the reprojection error between points is minimized, our refinement is expressed as a 6-DoF non-linear minimization problem, in which the Euclidean distance between points is replaced by a distance between ellipses, expressed as

\begin{equation}
    \hat{R}, \hat{t} = \argmin_{R, t} \sum_{j=1}^{N_{obj}} \Delta(\mathcal{E}_j, P Q_j^* P^T)^2,
\end{equation}
where $P = K [R|t]$ with $K$ being the calibration matrix of the camera and $[R|t]$ are its extrinsic parameters which are optimized. $\mathcal{E}_j$ is the j-th ellipse detected in the image, while $Q_j$ is the given dual-form matrix of the corresponding ellipsoid and $\Delta(\cdot)$ is the chosen ellipse-ellipse metric.

\subsection{Initial pose estimation and data association}
\label{subsec:init_pose_and_data_assoc}

While an initial camera pose can be obtained by different means (external sensor, odometry, pose prior), we rely on a slightly modified version of the method presented in~\cite{gaudilliere}, where the pose is solved inside a RANSAC loop which enumerates the different mapping possibilities, constrained by the objects labels.
At each iteration, a minimal set of three ellipse/ellipsoid pairs are selected and the camera pose is computed with the Perspective-3-Point (P3P) algorithm on the ellipses and ellipsoids centers. With only three points, P3P provides four potential solutions, from which the best one is selected based on the overlap (IoU) between the detected ellipses and the reprojected ones. The same IoU criteria is used to select the best camera pose from all the combinations evaluated in the RANSAC.
When only two objects are detected in the image, P3P is replaced by the original Perspective-2-Ellipsoid (P2E~\cite{gaudilliere}), which is able to compute the camera pose from only two ellipse/ellipsoid pairs, but under the additional assumption that the camera roll is null. In both cases, the camera pose is estimated by aligning the projection rays passing through the ellipsoids centers with the back-projection rays going through the detected ellipses centers, which is only approximate.
Finally, the pairs of ellipse/ellipsoid with a minimum projection overlap of 0.2 are selected as inliers and used in the optimization.

\subsection{Uncertainty estimation}



Inspired by the work of Gal~\cite{what_uncertainties} on aleatoric uncertainty and similarly to the work of Dong~\etal\cite{dong_isler} on object ellipsoid estimation, we propose a practically efficient method for estimating a predictive uncertainty on the detected ellipses. We then leverage it to weight the contribution of each object in the pose estimation.
However, instead of independently modelling each regressed parameter as a univariate Gaussian distribution, we predict a global measure of uncertainty representing the geometric quality of the regressed ellipse.
For that, we model the sampling-based distance ($\Delta$) used to train the ellipse prediction network~\cite{3DV} as a univariate Gaussian distribution. The minimisation objective becomes



\begin{equation}
    \mathcal{L}_{unc} = \frac{1}{2} \sigma^{-2} \Delta(\mathcal{E}_{pred}, \mathcal{E}_{gt}) + \frac{1}{2} \log{\sigma^2}.
    \label{equ:uncertainty_loss}
\end{equation}
In practice, we train the network to predict the log variance, $\alpha = \log{\sigma^2}$ to avoid numerical instability when $\sigma$ is small.
In the left part of the loss, $\sigma$ acts as a loss attenuation, whereas the right part is a regularizer term which should avoid predicting an infinite uncertainty. This loss attenuation is self-learned and allows the network to reduce its loss even in the hardest cases, where the network is not able to predict an ellipse close to the ground truth. To predict this additional value, we simply added a fully connected layer at the end of the Multi-Layer Perceptron (MLP) part of the network.
These predicted uncertainties $\sigma_j$ are then directly integrated in the pose optimization, such that
\begin{equation}
    \hat{R}, \hat{t} = \argmin_{R, t} \sum_{j=1}^{N_{obj}} \sigma_j^{-1} \Delta(\mathcal{E}_j, P Q_j^* P^T).
    \label{equ:uncertainty_weighting}
\end{equation}

\section{Experiments and Analysis}
\label{sec:experiments_camera_pose_refinement}

\subsection{Implementation details}

We performed all the optimizations in this work using the \textit{Broyden-Fletcher-Goldfarb-Shanno} algorithm (\textit{BFGS})~\cite{bfgs}.
For the level set metric, we tested different numbers of sampling points, starting from 1600 (40 azimuths and 40 distances). We reduced this number until 24 (6 azimuths and 4 distances) without degrading the accuracy and with an noticeable computation speedup.

\subsection{Ellipses alignment for camera pose refinement}

We evaluate the estimated camera poses obtained with the different ellipse-ellipse metrics on the 7-Scenes dataset~\cite{7-Scenes} (\textit{Chess} scene).
This scene is composed of 11 objects from 7 categories. We used the sequences 2, 3 and 5 for evaluation and sequences 1, 4 and 6 to build the scene model and train the object detection and ellipse prediction networks. Each sequence includes 1000 frames. We created the scene ellipsoidal model using the method from~\cite{rubino_ellipsoid_reconstruction}, with only a few manual annotations (a bounding box for each object in at least three images).
We fine-tuned Faster R-CNN and followed~\cite{3DV} to obtain 3D-aware elliptic detections of the objects.
The initial camera pose estimate and data associations are obtained in a RANSAC loop with either P2E or P3P, as explained in subsection~\ref{subsec:init_pose_and_data_assoc}.

\begin{figure*}
    \centering
    \includegraphics[width=\linewidth]{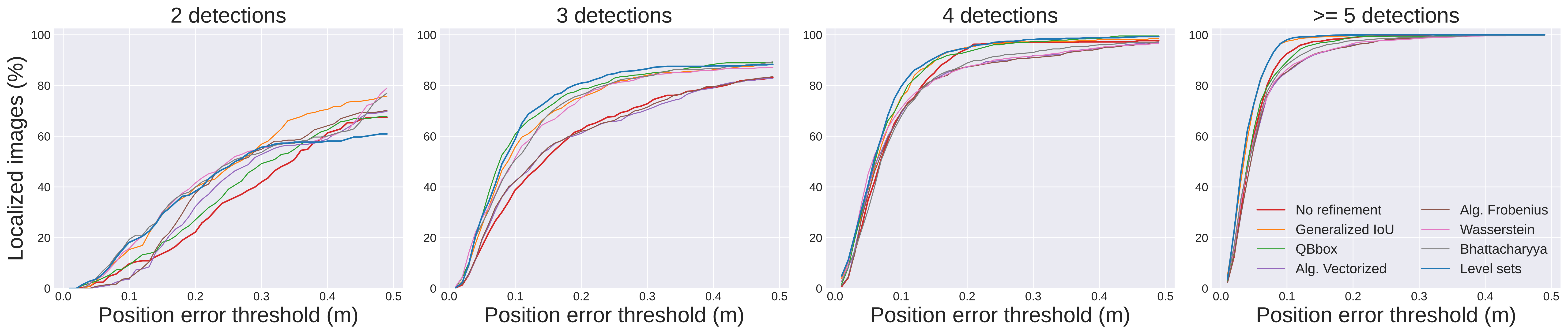}
    \caption{Percentage of correctly localized images w.r.t. an error threshold on the position. The images are split by the number of objects used in the pose optimization. The results obtained w.r.t an error threshold on the orientation are very similar.}
    \label{fig:chess_results}
\end{figure*}

\begin{figure}[ht]
    \centering
    \includegraphics[width=\linewidth]{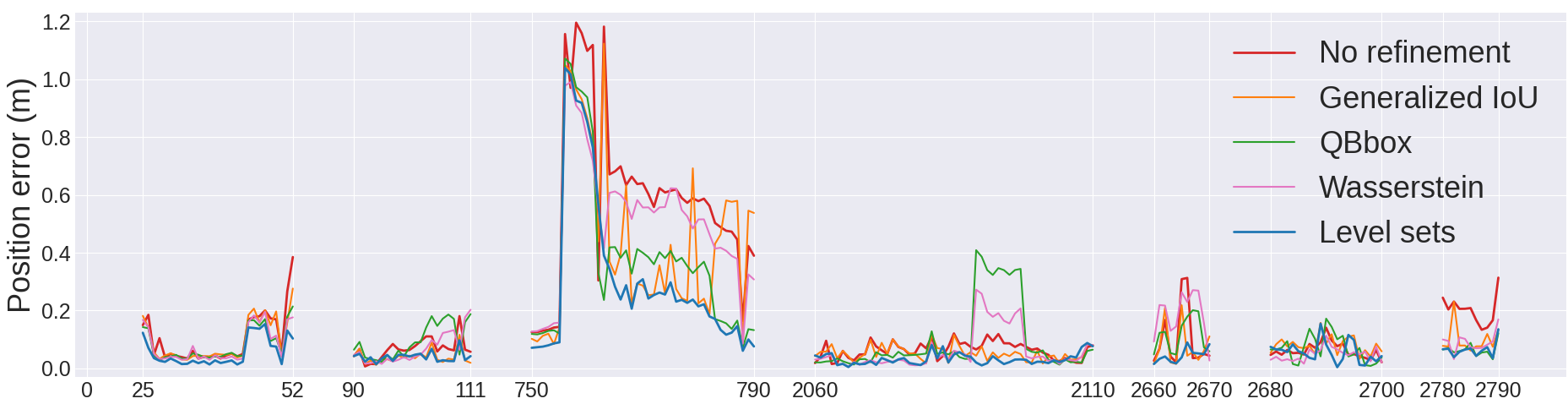}
    \caption{Position error obtained on the test frames (horiz. axis).}
    \label{fig:comparison_metrics_pos_errors_curves}
\end{figure}

Figure~\ref{fig:chess_results} shows the percentage of correctly localized frames over the test sequences with respect to an error threshold on the camera position. 
We separated the cases by the number of objects detected in the images and used in the optimization.
We can first notice that our pose estimation method, including the refinement step, clearly outperforms the direct closed-form solution~\cite{gaudilliere} (named "No refinement"). In terms of metrics used in the optimization, we can see that the level set one achieves the best results. The generalized IoU has also good performance in the case of four and more objects, but is clearly under-performing in images with three objects. On the contrary, QBbox performs well in the cases with three and four objects, but much less well than level set and GIoU starting from five objects.
We identify the test frames which produce a significant difference between the costs in Figure~\ref{fig:comparison_metrics_pos_errors_curves} and analyze them more precisely in the next section. The pose optimization requires a mean time of \SI{0.26}{\second} with the level set metric, faster than the two other best performing metrics GIoU (\SI{0.336}{\second}) and QBbox (\SI{0.313}{\second}).

\subsection{Costs behaviour analysis}
\label{subsec:costs_behaviours_analysis}

\begin{figure}
    \centering
    \includegraphics[width=\linewidth]{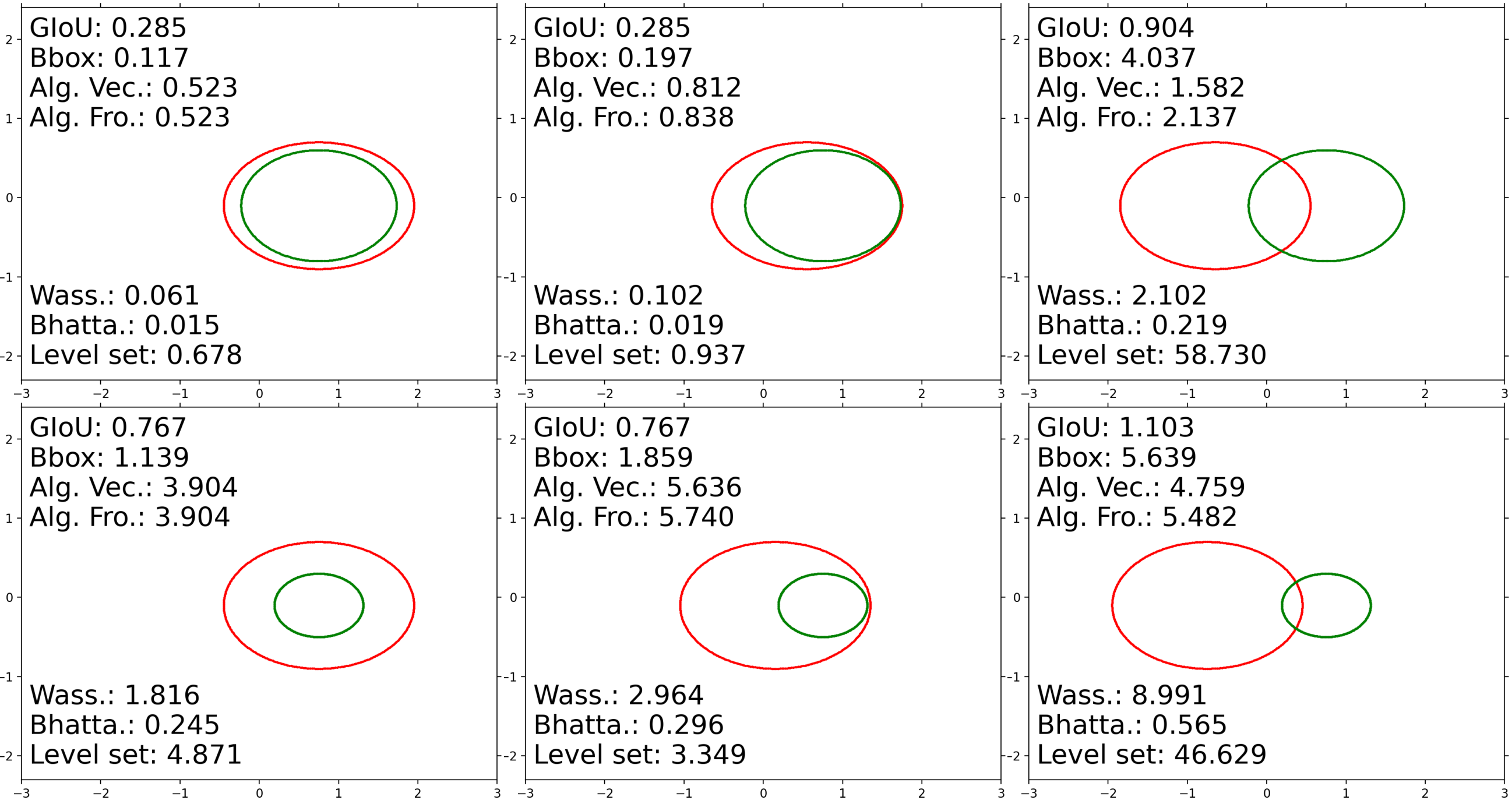}
    \caption{Ellipse-ellipse costs between a detection ellipse (\textit{green}) and a projection ellipse (\textit{red}).}
    \label{fig:animation_costs}
\end{figure}

In this section, we analyse the behaviour of the level set metric and highlight two nice properties. The bottom row of Figure~\ref{fig:animation_costs} illustrates the case of a detected ellipse (green) significantly smaller than the projected one (red) and, comparing the first and second columns, we can notice that all the metrics favour an alignment of the centers except the generalized IoU, whose cost value remains constant, and the level set metric which results in a lower cost in case of tangency. This behaviour of the level set metric is actually linked to the relative size between the detection and projection ellipses, favoring centers alignment when the sizes are similar (top row) and tangency as sizes differ (bottom row). More illustrations are available in the attached video. This behaviour proves to be particularly beneficial in the case of partially visible objects, for which aligning the centers does not make much sense and where tangency is more desirable. This is detailed in the next sub-section~\ref{subsubsec:partially_visible_objects}.

The third column of Figure~\ref{fig:animation_costs} also shows that the level set cost increases much faster than the other metrics when the ellipses become distant. This induces that this particular metric highly penalizes distant ellipses, which proves to be more efficient. It will be illustrated in sub-section~\ref{subsubsec:small_objects}.


\subsubsection{Partially visible objects}
\label{subsubsec:partially_visible_objects}
\begin{figure}
    \centering
    \includegraphics[width=\linewidth]{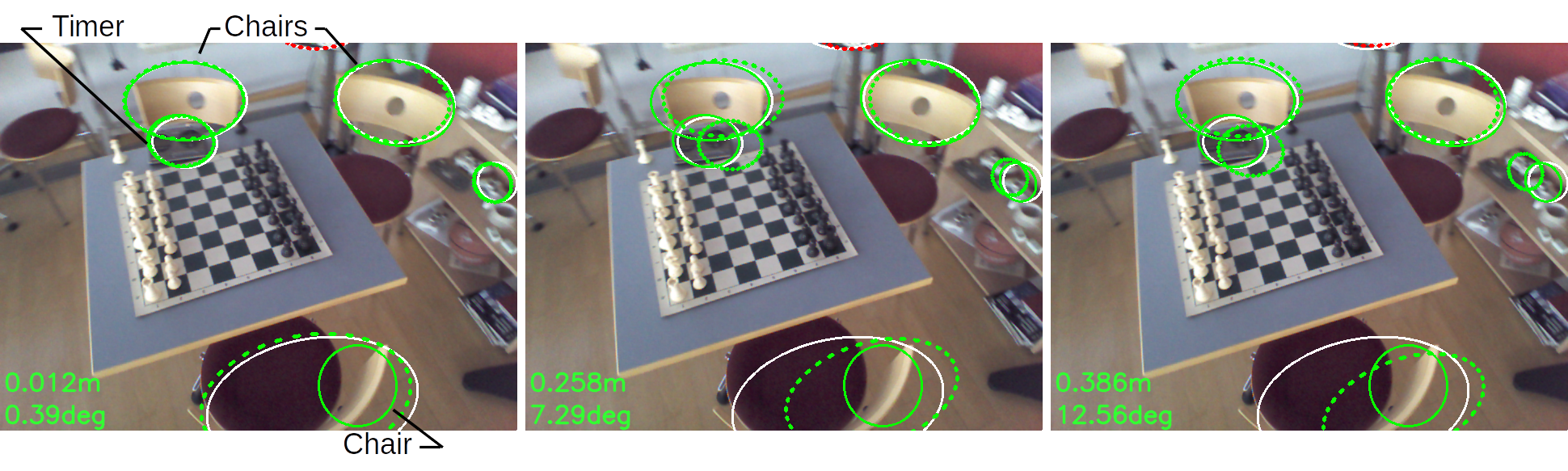}
    \caption{Estimated pose using level set (left), Wasserstein (center) and QBbox (right). The green solid-line ellipses are the detections and the dashed ones are the objects projections obtained with the estimated camera pose. The green ellipses are used in the pose optimization, the red ones (if any) do not match any detection and the white ones are the objects projections obtained with the ground truth camera pose. The position and orientation errors are written in green.}
    \label{fig:results_chess_frames_comparison_frames_2090}
\end{figure}

Figure~\ref{fig:results_chess_frames_comparison_frames_2090} illustrates the case of poor-quality detection for a partially visible object (the bottom chair is slightly outside the image and seen from an uncommon viewpoint) and shows the benefits of using the level set metric.

We can first notice that the Wasserstein distance tends to align the center of the bad chair detection with the projection of its model, at the expense of degrading the alignment of the well detected objects. Actually, the Wasserstein cost sums a distance between centers and a difference in shapes (see Equation~\ref{equ:Wasserstein}). This explains why the alignment of the bottom chair (with a large size difference) is prioritized compared the top chair and the timer (which have correct sizes).
Minimizing the distance between on-image bounding boxes (QBbox) is also problematic for this frame.
In fact, QBbox is efficient to deal with poor-quality detections, but only when the missing part is outside the image. In this case, however, the non-detected part of the object is still situated inside the image and the bad detection is also due to the uncommon viewpoint on the object. 
This induces a high cost for the bottom chair and results in a bad estimated pose.
On the contrary, we showed in Figure~\ref{fig:animation_costs} that the level set metric favours tangency when the detection is significantly smaller than the projection. This results in a much better alignment for the well detected objects and a noticeably better pose accuracy.

\subsubsection{Near vs. far objects contributions}
\label{subsubsec:small_objects}

\begin{figure}[H]
    \centering
    \includegraphics[width=\linewidth]{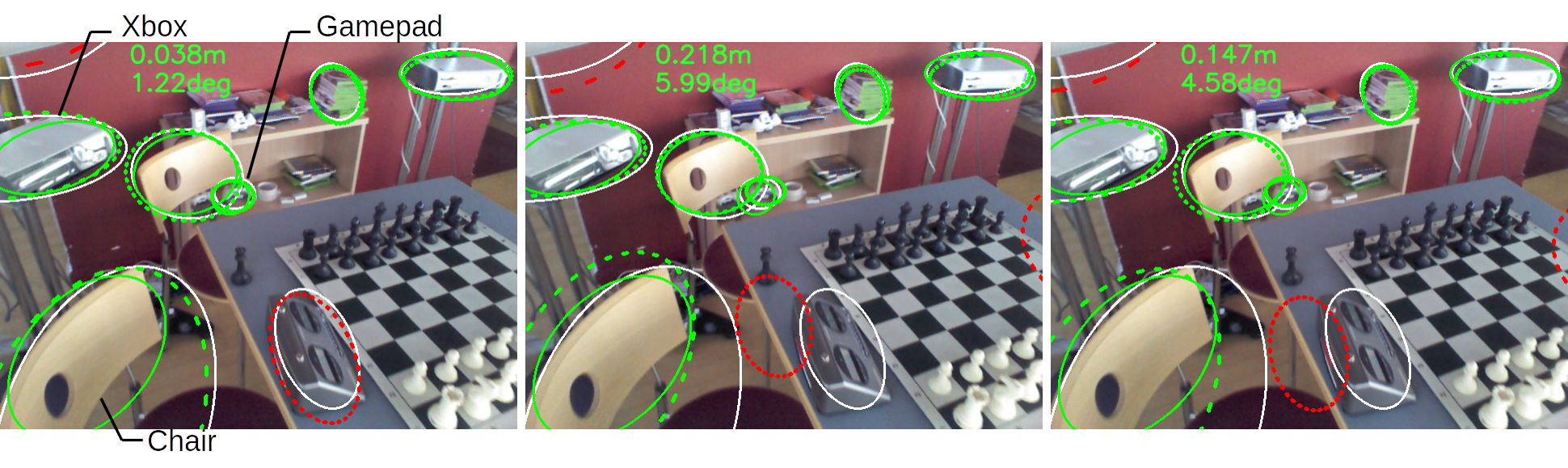}
    \caption{Estimated pose using level set (left), GIoU (center) and QBbox (right). The color code is the same as in figure~\ref{fig:results_chess_frames_comparison_frames_2090}.}
    \label{fig:results_chess_frames_comparison_frames_2665}
\end{figure}


Figure~\ref{fig:results_chess_frames_comparison_frames_2665} compares the results obtained with level set, GIoU and QBBox. Although this frame contains partially visible objects (the bottom chair and the left Xbox), these are not problematic for the selected metrics. Indeed, GIoU does not enforce centers alignment for a smaller detection ellipse (the cost is flat) and QBbox is able to correctly handle the poorly detected objects because the missing parts are situated outside the image. The challenge, here, is more related to adjusting the contribution between objects, especially between the near and far ones.
Far objects are more subject to misalignment between their detection and projection ellipses when the pose estimate is erroneous. Consequently, they constitute good anchors for the pose estimation and a strong misalignment should be highly penalized.


Absence of normalization, such as with QBbox, naturally puts more importance on the large objects in the image (i.e., near and/or large objects in the scene). For its part, the generalized IoU is naturally normalized, and thus, tends to equalize the objects contributions. Also, as shown in the third column in Figure~\ref{fig:animation_costs}, this cost increases very slowly when the ellipses move away from each other.
On the contrary, the level set metric is also naturally normalized by the chosen embedding function (the ellipse contour always corresponds to the level curve of value 1), but produces a high cost for distant ellipses (see Figure~\ref{fig:animation_costs}). This metric therefore takes more advantage of the far objects.
Indeed, we can observe a better alignment between detection and projection ellipses for the gamepad and a better pose accuracy using the level set metric in Figure~\ref{fig:results_chess_frames_comparison_frames_2665}.

\subsection{Convergence analysis}
\begin{figure}
    \centering
    \includegraphics[width=\linewidth]{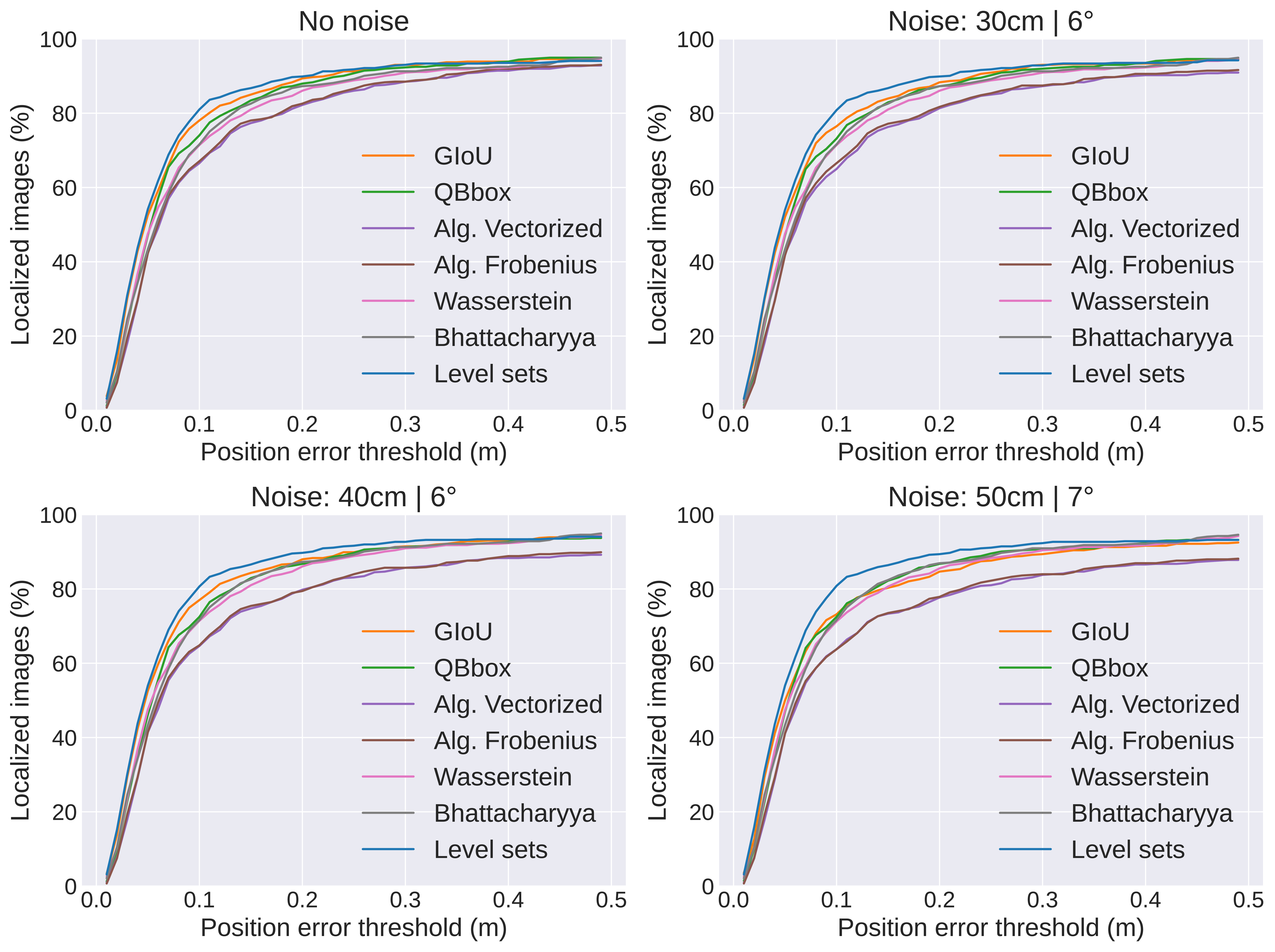}
    \caption{Percentage of correctly localized images w.r.t. a position error threshold for an increasing level of noise added to the initial pose estimate.}
    \label{fig:noisy_pose_init}
\end{figure}


We randomly simulated noisy initial camera poses to evaluate the convergence basins of the different metrics. The results are available in Figure~\ref{fig:noisy_pose_init} for different levels of noise on the camera position and orientation. We can notice that the level set metric is particularly robust. Wasserstein and Bhattacharrya are only slightly affected by noise, but they globally perform less well. On the contrary, QBbox, GIou and the algebraic distances are more strongly affected by a noisy initial camera pose.

\subsection{Uncertainty-driven pose estimation}

\begin{figure}[ht]
    \centering
    \includegraphics[width=\linewidth]{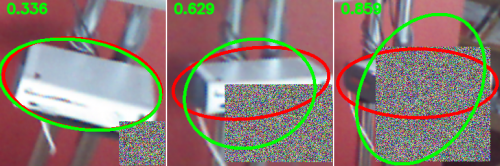}
    \caption{Predicted (green) and ground truth (red) ellipses for an increasing level of occlusion. The predicted uncertainties are written in the top-left corners.}
    \label{fig:uncertainty_correlation}
\end{figure}

\begin{figure}[ht]
    \centering
    \includegraphics[width=1\linewidth]{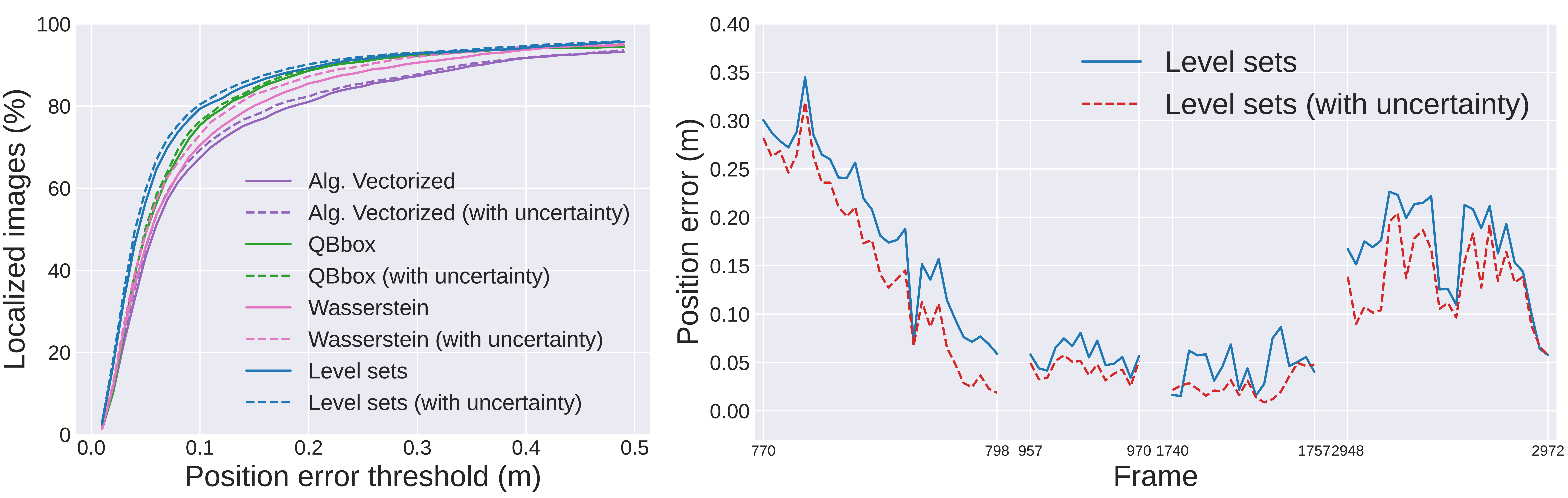}
    \caption{Comparison between the standard version and the uncertainty-driven pose optimization.}
    \label{fig:uncertainty_results_curves}
\end{figure}

\begin{figure}[ht]
    \centering
    \includegraphics[width=\linewidth]{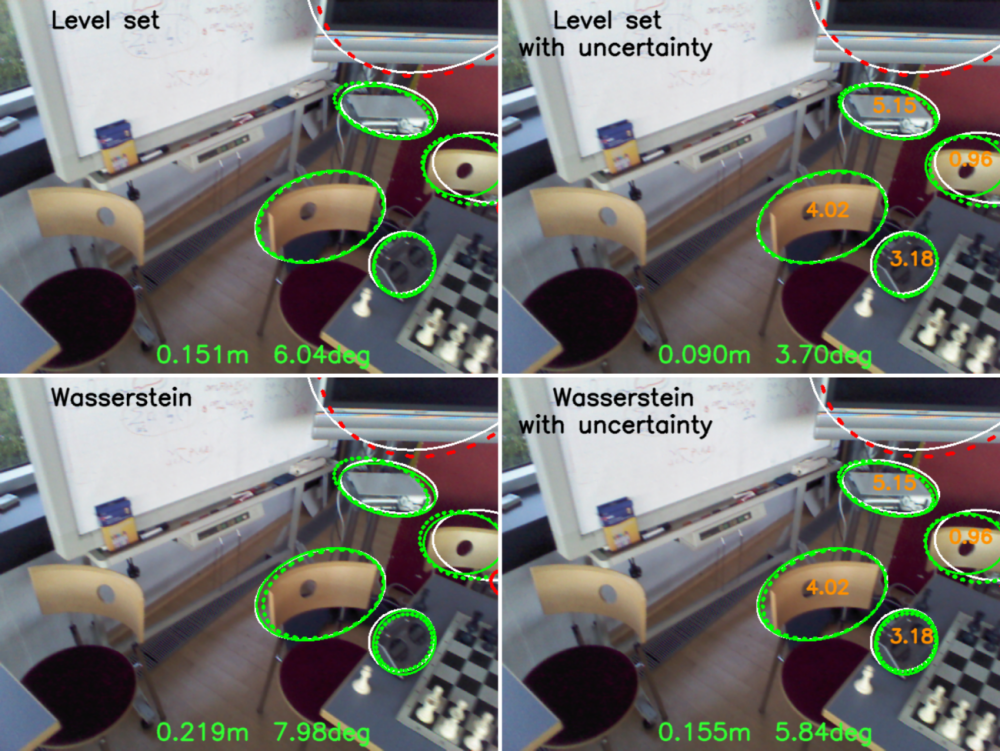}
    \caption{Comparison between the standard (left) and the uncertainty-driven estimation (right). The level set metric is used in the top row and the Wasserstein distance in the bottom row. The color code is the same as in Figure~\ref{fig:results_chess_frames_comparison_frames_2090}. Additionally, the objects weighting are shown in orange.}
    \label{fig:uncertainty_results_images}
\end{figure}

We first show the reliability of the predicted uncertainty on crop images of the objects. We added an increasing level of occlusion to the image to simulate a low-quality ellipse prediction. 
As expected, the estimated uncertainty increases with the occlusion ratio (see Figure~\ref{fig:uncertainty_correlation}).
We then evaluate the benefits offered by the uncertainty guidance in the camera pose refinement. Globally, the weighting proposed in equation~\ref{equ:uncertainty_weighting} improves the pose accuracy (see Figure~\ref{fig:uncertainty_results_curves}). Some particularly interesting cases are shown in Figure~\ref{fig:uncertainty_results_images}.
We can first notice that the estimated pose (without uncertainty) is better using the level set metric compared to Wasserstein. This can be explained by the right chair which is slightly outside the image and whose detection is smaller than expected. As explained in section~\ref{subsubsec:partially_visible_objects}, the level set metric handles this case better than the Wasserstein distance which pushes more towards centers alignment. Additionally weighting the objects costs by confidence still further improves the estimated pose for both metrics. We can observe that projections of the right chair is more loosely constrained by the detection, which also allows the more confident objects to reach a better alignment (the Xbox and the left chair).
Figure~\ref{fig:uncertainty_results_curves} shows the global improvement of using uncertainty on the entire test sequence. In particular, we can notice that the benefit obtained for the Wasserstein distance is slightly larger than for the level set or QBbox metrics, probably because these two metrics naturally deal better with partially visible objects.

\section{Conclusion}

We proposed a camera pose estimation method based on objects, which aims at aligning their detections in the image with the projection of their ellipsoidal 3D models. We showed that establishing a cost between ellipses is not as straightforward as it is for points and that a metric based on level sets has good convergence properties. In particular, it helps to deal with the challenging case of partially visible objects.
Finally, we proposed a practical method for predicting uncertainty in a deep neural network, which reflects the quality of the predicted objects ellipses and showed how the weighting of each object contribution can help to improve the accuracy of the refined pose.

\bibliographystyle{IEEEtran}
\bibliography{pose}




\end{document}